\documentclass[10pt,twocolumn,letterpaper]{article}

\usepackage[utf8]{inputenc}
\usepackage{cvpr}
\usepackage{times}
\usepackage{epsfig}
\usepackage{graphicx}
\usepackage{amsmath,amsfonts,amssymb,amsthm}
\usepackage{color}
\usepackage{amsfonts}
\usepackage{caption}
\usepackage{subcaption}
\usepackage{algorithm}
\usepackage[noend]{algpseudocode}
\usepackage{xspace}
\usepackage{ifthen}
\usepackage{cite}
\usepackage{array}
\usepackage{soul}
\usepackage[pagebackref=true,breaklinks=true,letterpaper=true,colorlinks,bookmarks=false]{hyperref}

\def\cvprPaperID{1448}

\cvprfinalcopy

\graphicspath{ {gfx/} }

\newcommand{\bs}{\mathbf{s}}

\newcommand{\bp}{\mathbf{p}}

\newcommand{\bc}{\mathbf{c}}

\newcommand{\bpi}{\boldsymbol{\pi}}

\newcommand{\cR}{\mathcal{R}}

\newcommand{\cL}{\mathcal{L}}

\newcommand{\cP}{\mathcal{P}}

\newcommand{\cF}{\mathcal{F}}

\newcommand{\figref}[1]{\Fig~\ref{#1}}
\newcommand{\secref}[1]{Section~\ref{#1}}

\renewcommand{\eqref}[1]{Eq.~\ref{#1}}
\newcommand{\tabref}[1]{Table~\ref{#1}}

\makeatletter
\DeclareRobustCommand\onedot{\futurelet\@let@token\@onedot}
\def\@onedot{\ifx\@let@token.\else.\null\fi\xspace}
\def\eg{e.g\onedot} 
\def\ie{i.e\onedot}

\def\wrt{wrt\onedot}

\def\etal{et~al\onedot} 
\def\Fig{Fig\onedot}   
\makeatother

\ifcvprfinal\fi

\begin{document}

\title{Semantic Instance Annotation of Street Scenes by 3D to 2D Label Transfer}

\author{Jun Xie$^1$ \qquad Martin Kiefel$^2$ \qquad Ming-Ting Sun$^1$ \qquad Andreas Geiger$^2$\\
$^1$University of Washington \qquad $^2$MPI for Intelligent Systems T\"ubingen\\
{\tt\small \{junx,mts\}@uw.edu} \qquad {\tt\small \{martin.kiefel,andreas.geiger\}@tue.mpg.de}
}

\setlength\arraycolsep{2pt}

\maketitle

\begin{abstract}
Semantic annotations are vital for training models for object recognition, semantic segmentation or scene understanding. Unfortunately, pixelwise annotation of images at very large scale is labor-intensive and only little labeled data is available, particularly at instance level and for street scenes. In this paper, we propose to tackle this problem by lifting the semantic instance labeling task from 2D into 3D. Given reconstructions from stereo or laser data, we annotate static 3D scene elements with rough bounding primitives and develop a model which transfers this information into the image domain. We leverage our method to obtain 2D labels for a novel suburban video dataset which we have collected, resulting in 400k semantic and instance image annotations. A comparison of our method to state-of-the-art label transfer baselines reveals that 3D information enables more efficient annotation while at the same time resulting in improved accuracy and time-coherent labels.
\end{abstract}

\section{Introduction}

\begin{figure}[t]
\centering
\includegraphics[width=0.8\linewidth]{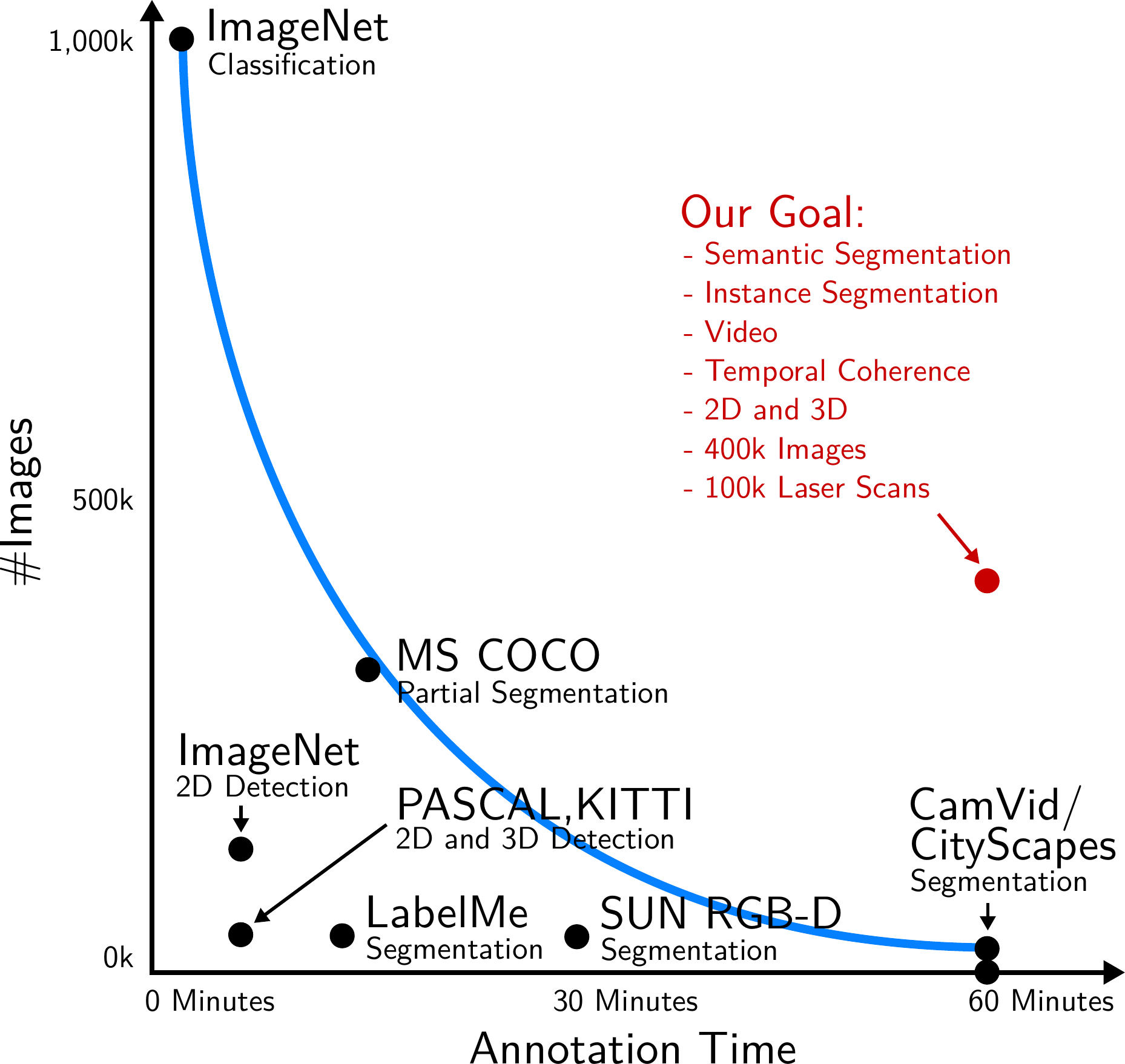}
\caption{{\bf The Curse of Dataset Annotation.}}
\label{fig:curse_annotation}
\vspace{-0.5cm}
\end{figure}

The revolutionary success of high-capacity deep learning architectures \cite{Krizhevsky2012NIPS,Zhu2015CVPR,Long2015CVPR} may flag the beginning of a paradigm shift in computer vision. Rather than developing methods for solving a certain task, future research could be directed towards teaching a ``universal program'' (\eg, a deep network) a mapping from input to output space.
One fundamental question arising in this context is how the required ground truth labels for training these models can be generated at very large scales (\ie, $>100$k images).
While for some tasks large annotated datasets are already available today (\eg, image classification \cite{Russakovsky2014ARXIV}), other tasks such as semantic segmentation of street scenes lack this information as human annotation is labor-intensive. We refer to this phenomenon as the {\bf curse of dataset annotation} (\figref{fig:curse_annotation}).

One option to circumvent this problem is to exploit auxiliary tasks for which large annotated datasets are available.
While generalization to the target domain can be achieved to some extent, discriminative cues which solve the auxiliary problem will dominate the learned representation \cite{Zhou2015ICLR}.
A second option is the creation of synthetic datasets.
Unfortunately, our community still lacks rich generative image formation models which are able to produce realistic and diverse imagery from the true underlying distribution of the 3D world we live in.
In this paper, we therefore propose an alternative approach which leverages additional 3D information to simplify the 2D annotation task.

Recently, applications such as autonomous cars and humanoid robots have attracted significant attention. For research in these applications, a street view video dataset with dense semantic labels will be very useful.
Motivated by those needs, our work focuses on the challenging task of semantic and instance video annotation of street scenes
for which pixelwise labeling requires up to $60$ minutes per image for a human annotator as acknowledged in \cite{Badrinarayanan2010CVPR}. Inspired by the easy usage of 3D modeling tools (Blender, SketchUp) we propose to annotate scenes directly in 3D and then transfer this knowledge back into the image domain. The required 3D information can be obtained from various sources including structure-from-motion (SfM), stereo or laser scanners. This approach has several advantages over labeling in 2D: First, objects often project into several images of the video sequence, thus lowering annotation efforts considerably. Further, the obtained 2D instance annotations are temporally coherent as they are associated with a single object in 3D. And finally, our 3D annotations might be useful by themselves for reasoning in 3D \cite{Zhang2013ICCV,Geiger2015GCPR} or to enrich 2D annotations with approximate 3D geometry.

\begin{figure}[t]
\begin{subfigure}{.48\linewidth}
\includegraphics[width=\linewidth]{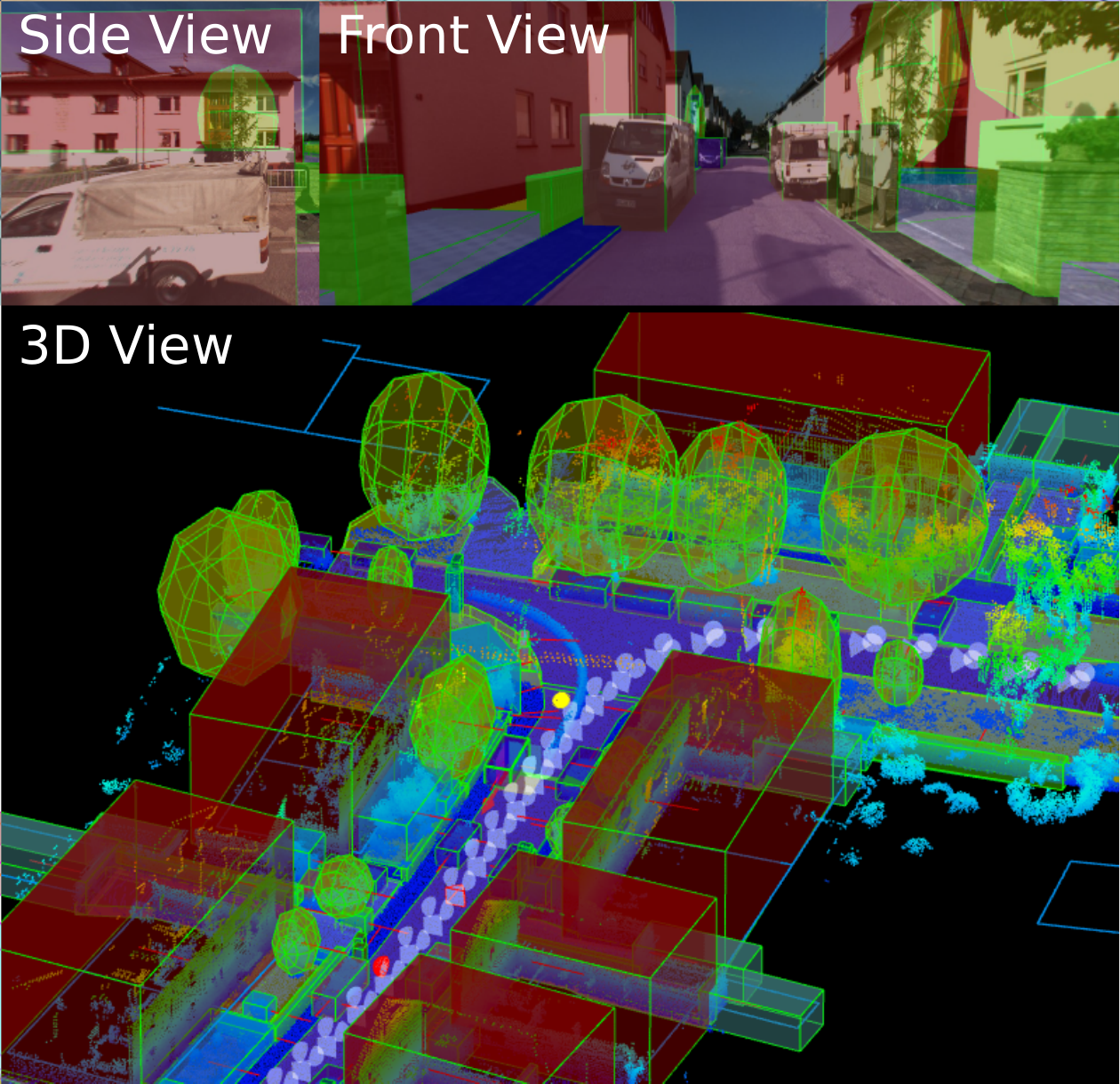}
\caption{3D Annotation}
\label{fig:illustration_gui}
\end{subfigure}%
\hfill
\begin{subfigure}{.48\linewidth}
\includegraphics[width=\linewidth]{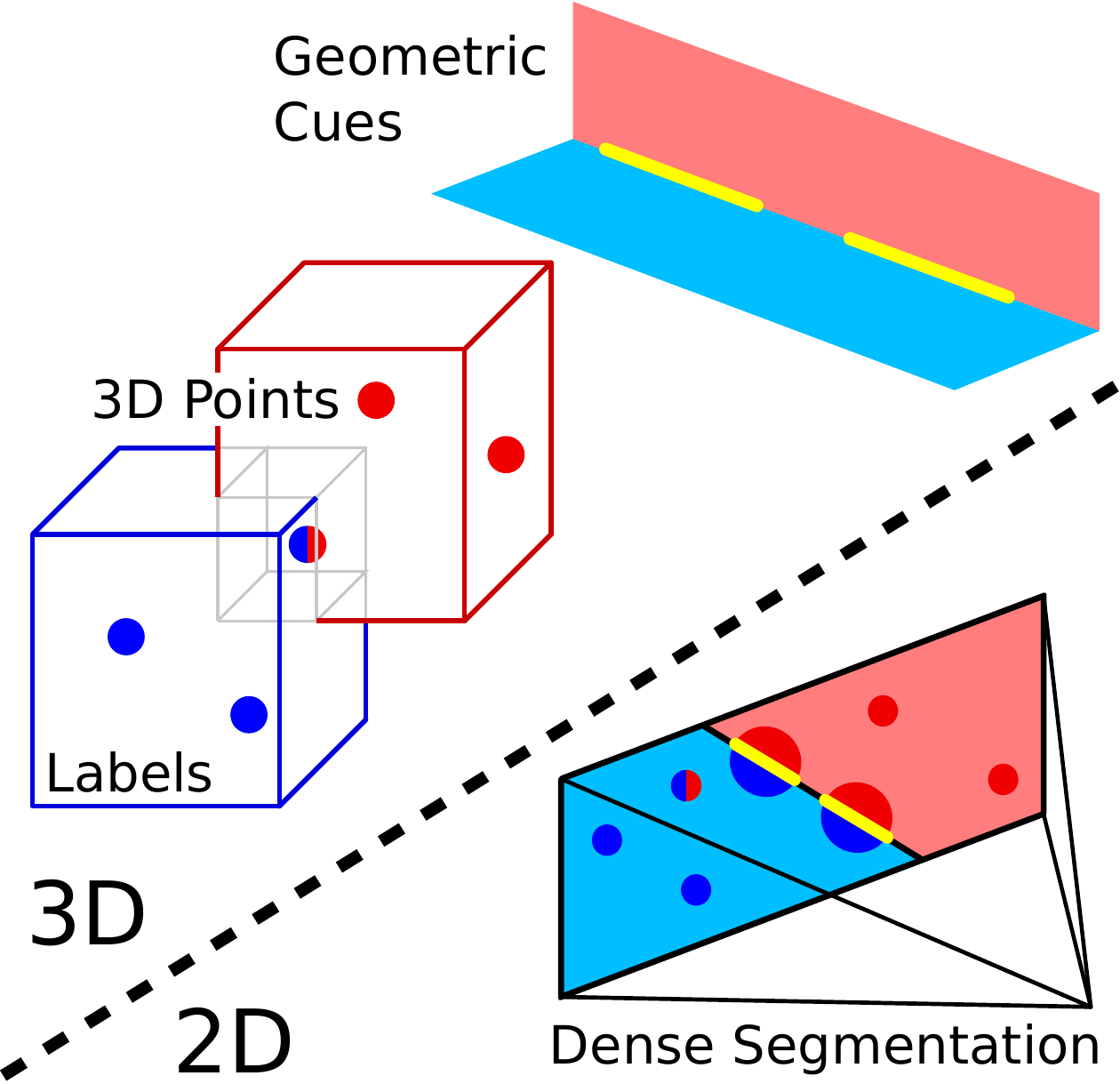}
\caption{Label Transfer}
\label{fig:illustration_method}
\end{subfigure}\\
\begin{subfigure}{\linewidth}
\includegraphics[width=\linewidth]{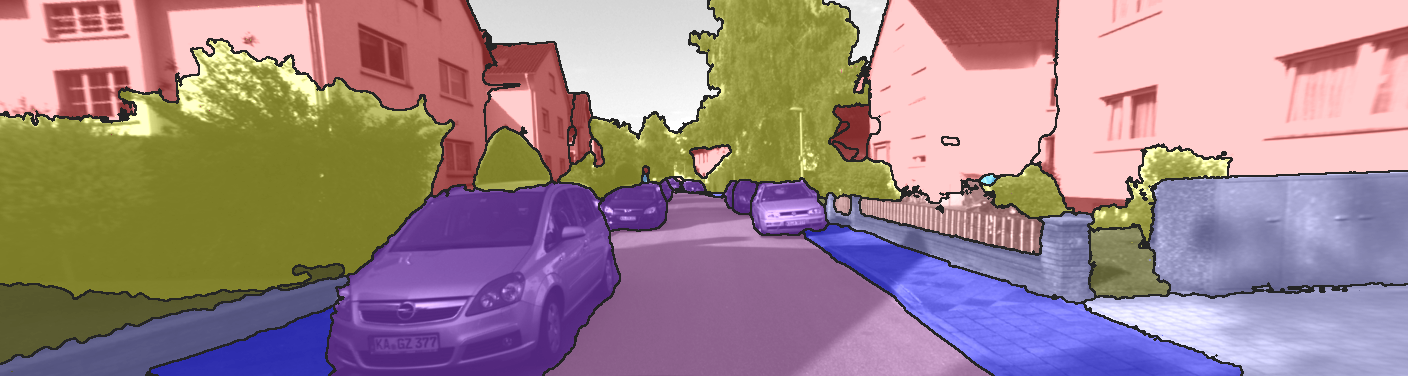}
\caption{2D Annotation}
\label{fig:illustration_results}
\end{subfigure}%
\caption{{\bf 3D to 2D Label Transfer:} (\subref{fig:illustration_gui}) We annotate all objects in 3D using bounding primitives. (\subref{fig:illustration_method}) Our model then transfers this information into 2D by jointly reasoning about 3D geometric cues, sparse 3D points, as well as image pixels. (\subref{fig:illustration_results}) This allows us to infer temporally consistent semantic instance annotations for every frame in the video.}
\label{fig:illustration}
\vspace{-0.5cm}
\end{figure}

Unfortunately, obtaining dense and accurate 2D labels from sparse noisy point clouds and coarse 3D annotations is a challenging task by itself.
Towards solving this problem, we propose a non-local multi-field CRF model which reasons jointly about semantic and instance labels of all 3D points and all pixels in the image as illustrated in \figref{fig:illustration}. This approach offers several advantages over methods which reason purely in 2D \cite{Badrinarayanan2014IJCV,Vijayanarasimhan2012ECCV}: Occluders and occludees which exhibit complex boundaries when projected onto the image plane (\eg, tree in front of a building) are often easier to separate in 3D. Besides, our approach is not affected by missing labels due to occlusions or drift in optical flow.
Further, our model allows to specify a tractable semantic instance loss for principled and efficient end-to-end parameter learning. And finally, the probabilistic nature of our model allows for estimating label uncertainties which can be used to increase label accuracy when only a subset of the pixels require a label.
In summary, we make the following two contributions in this paper:
\begin{itemize}
\item We present a novel geo-registered dataset of suburban scenes recorded by a moving platform. The dataset comprises over $400$k images and over $100$k laser scans, and we provide semantic 3D annotations for all static scene elements.
\item We propose a method which is able to transfer these labels from 3D into 2D, yielding pixelwise semantic instance annotations. We demonstrate the potential of our approach in ablation studies and with respect to several 2D and 3D baselines.
\end{itemize}

We make our code, dataset and annotations publicly available at \url{http://www.cvlibs.net/projects/label_transfer}.

\section{Related Work}

In this section, we first review semi-supervised video annotation methods, followed by an overview over existing semantic and instance segmentation datasets.

\vspace{-0.4cm}
\paragraph{Methods:}

Compared to annotating individual images
\cite{Xu2014CVPR,Guillaumin2014IJCV,Liu2011PAMI},
video sequences offer the advantage of temporal coherence between adjacent frames. Label propagation techniques exploit this fact by transferring labels from a sparse set of annotated keyframes to all unlabeled frames based on color and motion information. While in some works a single foreground object is assumed
\cite{Jain2014ECCV,Tsai2012IJCV},
here we focus on methods which can handle multiple object categories.
Towards this goal, \cite{Badrinarayanan2010CVPR,Budvytis2010BMVC} proposed a coupled Bayesian network based on video epitomes and semantic regions to propagate label information between two annotated keyframes.
To better account for errors in label propagation, \cite{Nagaraja2012DAGM} proposed a hierarchy of local classifiers for this task and \cite{Badrinarayanan2014IJCV} leveraged a mixture-of-tree model for temporal association.
The problem of selecting the most promising key frames for annotation has been considered in \cite{Vijayanarasimhan2012ECCV}.

In contrast to the aforementioned methods which propagate labels in 2D, in this paper we propose to annotate directly in 3D and then project these annotations into the 2D domain. While this approach requires a source of 3D information (\eg, SfM, stereo, laser),
it is able to produce more accurate semantic and temporally consistent instance annotations. Further, our experiments indicate that annotation in 3D is more time efficient than labeling in 2D as scene elements can be separated more easily and often project into many images of the input video sequence.

There exists little work on 3D to 2D label transfer. A notable exception is the approach of Chen \etal \cite{Chen2014CVPRb}, where annotations from KITTI \cite{Geiger2013IJRR} as well as 3D car models are leveraged to infer separate figure-ground segmentations for all vehicles in the image. In  comparison, our approach reasons jointly about all objects in the scene and also handles categories for which CAD models or 3D point measurements are unavailable (\eg, ``Tree'', ``Sky'').
In the context of street view image segmentation, Xiao \etal \cite{Xiao2009ICCV} present a hybrid method where annotated 3D points from structure-from-motion are projected onto superpixels in the image and users interactively correct wrong predictions with 2D scribbles. However, as no occlusion reasoning is performed, their method can only be applied to scenes with little variations in depth (\eg, facades).
Other methods \cite{Brostow2008ECCV,Munoz2009CVPR,Munoz2012ECCV,Namin2015WACV,Martinovic2015CVPR} which model the interaction between image pixels and 3D points focus primarily on improving classification performance or efficiency by exploiting multiple input modalities while our goal is to transfer ambiguous 3D primitive labels to every pixel in the image.

\vspace{-0.4cm}
\paragraph{Datasets:}

While some datasets such as PASCAL VOC \cite{Everingham2010IJCV} or MS COCO \cite{Lin2014ECCV} provide semantic labels for a subset of pixels in the image, here we focus on datasets with dense semantic annotations. Most of these datasets provide only a small number  ($\sim1$k) of accurately annotated indoor\cite{Silberman2012ECCV} or outdoor \cite{Shotton2009IJCV,Gould2009ICCV} images. A notable exception is LabelMe \cite{Russell2008IJCV} with more than $10$k images labeled using crowdsourcing techniques. Compared to the smaller datasets, however, not all images are densely annotated, quality varies heavily amongst annotators, and polygons have been chosen over pixels as more efficient but less accurate representation.

A number of works have also considered the annotation of video sequences \cite{Brostow2009PRL,Xiao2013ICCV,Song2015CVPR}. In \cite{Xiao2013ICCV}, eight \mbox{RGB-D} sequences of indoor scenes have been manually annotated using an interactive tool which propagates 2D polygons from one frame to another. The recently proposed SUN \mbox{RGB-D} dataset \cite{Song2015CVPR} provides labeled 2D polygons as well as 3D cuboids for $10$k \mbox{RGB-D} images captured indoors. For street scenes, less annotated data is available \cite{Behley2012ICRA,Munoz2009CVPR,Munoz2012ECCV,Riemenschneider2014ECCV,Valentin2013CVPR}. While KITTI \cite{Geiger2012CVPR} provides semantic information only for a few object categories\footnote{\url{http://www.cvlibs.net/datasets/kitti/}}, CamVid \cite{Brostow2009PRL} offers pixel-accurate labels, but without instances and for a very limited number of frames.
Very recently, the Cityscapes dataset \cite{Cordts2016CVPR} has been proposed with $5$k manually annotated individual 2D images of street scenes\footnote{\url{http://www.cityscapes-dataset.net/}}. Our dataset differs from Cityscapes in that we provide temporally coherent semantic instance annotations at a much larger scale as well as omnidirectional imagery, 3D laser scans and 3D annotations which might also be directly useful for reasoning in 3D.
While \cite{Cordts2016CVPR} focuses on inner-city scenes, our dataset comprises mainly suburban areas, thus both datasets complement each other.

\section{Method}

In this work, we are interested in generating semantic instance annotations for urban scenes at large scale by transferring labels from sparse 3D point clouds into the images.
In particular, we focus on static scene elements which dominate suburban scenes. Dynamic objects could be handled via 3D models \cite{Menze2015CVPR,Chen2014CVPRb} but as our dataset comprises little dynamic objects we leave this extension for future work.
This section describes our data collection efforts, our 3D annotation process, as well as the proposed label transfer model.

\subsection{Data Collection}
\label{sec:data_collection}

For our data collection, we equipped a station wagon with one $180^\circ$ fisheye camera to each side and a $90^\circ$ perspective stereo camera (baseline $60$ cm) to the front. Furthermore, we mounted a Velodyne HDL-64E and a SICK LMS 200 laser scanning unit in pushbroom configuration on top of the roof. This setup is similar to the one used in KITTI \cite{Geiger2012CVPR,Geiger2013IJRR}, except that we gain a full $360^\circ$ field of view due to the additional fisheye cameras and the pushbroom laser scanner while KITTI only provides perspective images and Velodyne laser scans with a $26.8^\circ$ vertical field of view. Compared to omnidirectional camera systems \cite{Schoenbein2014IROS,Schoenbein2014ICRA} our setup benefits from increased resolution.
Approximate localization is provided by an IMU/GPS measurement unit.

Using this setup, we recorded several suburbs of a mid-size city corresponding to over $400$k images and $100$k laser scans.
We estimated all vehicle and camera poses using structure-from-motion \cite{Heng2013IROS}.
More specifically, we minimize 3D reprojection errors based on all feature matches while regularizing against the GPS solution. This results in accurate georegistered camera poses. While our label transfer approach does not assume geolocalization, geospatial information\footnote{\url{http://www.openstreetmap.org/}} can facilitate the 3D annotation task.

\subsection{Annotation}

We augmented our dataset with 3D annotations in the form of bounding primitives, \ie, we placed cuboids and ellipsoids around objects in 3D and assigned a semantic label to each of them. More specifically, we asked a group of annotators to tightly enclose the 3D points belonging to an object by the respective primitive. For this purpose, we developed a 3D annotation tool based on WebGL (see \figref{fig:illustration_gui}) which visualizes the colored point clouds (obtained by projecting the 3D points back onto multiple images), two camera views, and provides tools to facilitate navigation and annotation.
To enable efficient annotation, our primitives are rough approximations of the true object shapes and thus are allowed to overlap in 3D (see \figref{fig:illustration_method}).
For stuff categories (\eg, ``Road'', ``Sidewalk'', ``Grass'') we allow users to draw 2D polygons in bird's eye view which are then extruded into 3D to better approximate the shape and to facilitate annotation. Ambiguities are resolved using our label transfer method described in the following section. Annotating a single batch comprising 200 laser scans and 800 images required about 3 hours.
While the focus of this paper is on annotating static scene elements which cover the majority of pixels in general, our annotation GUI could be extended to a keyframe based dynamic 3D video annotation tool which visualizes point clouds and images over time akin to the annotation utility developed for labeling the KITTI dataset \cite{Geiger2012CVPR,Geiger2013IJRR}.

\subsection{Model}
\label{sec:label_transfer}

\begin{figure}[t]
\begin{subfigure}{.48\linewidth}
\includegraphics[width=\linewidth]{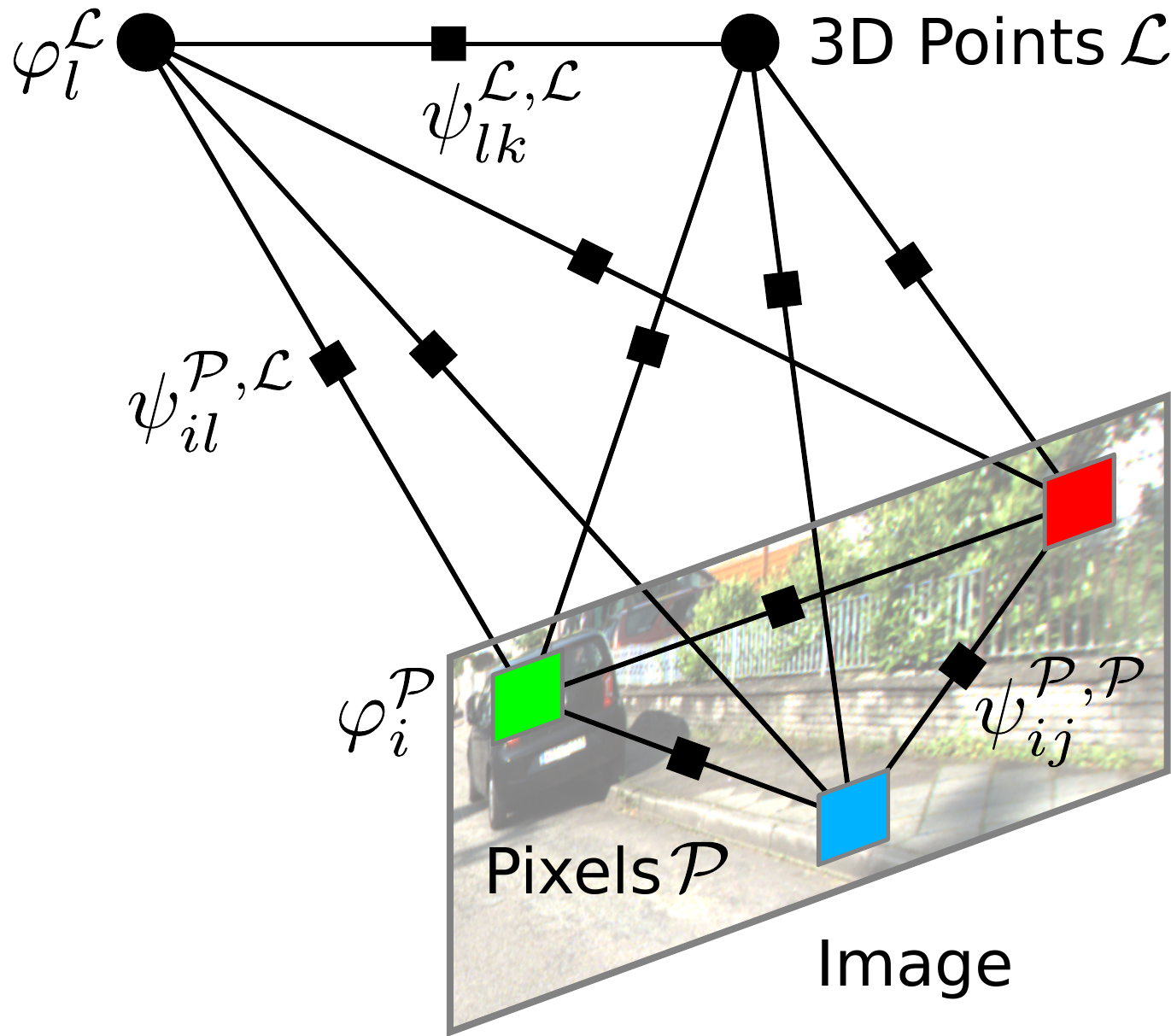}
\caption{Graphical Model}
\label{fig:model_dense_crf}
\end{subfigure}%
\hfill
\begin{subfigure}{.48\linewidth}
\includegraphics[width=\linewidth]{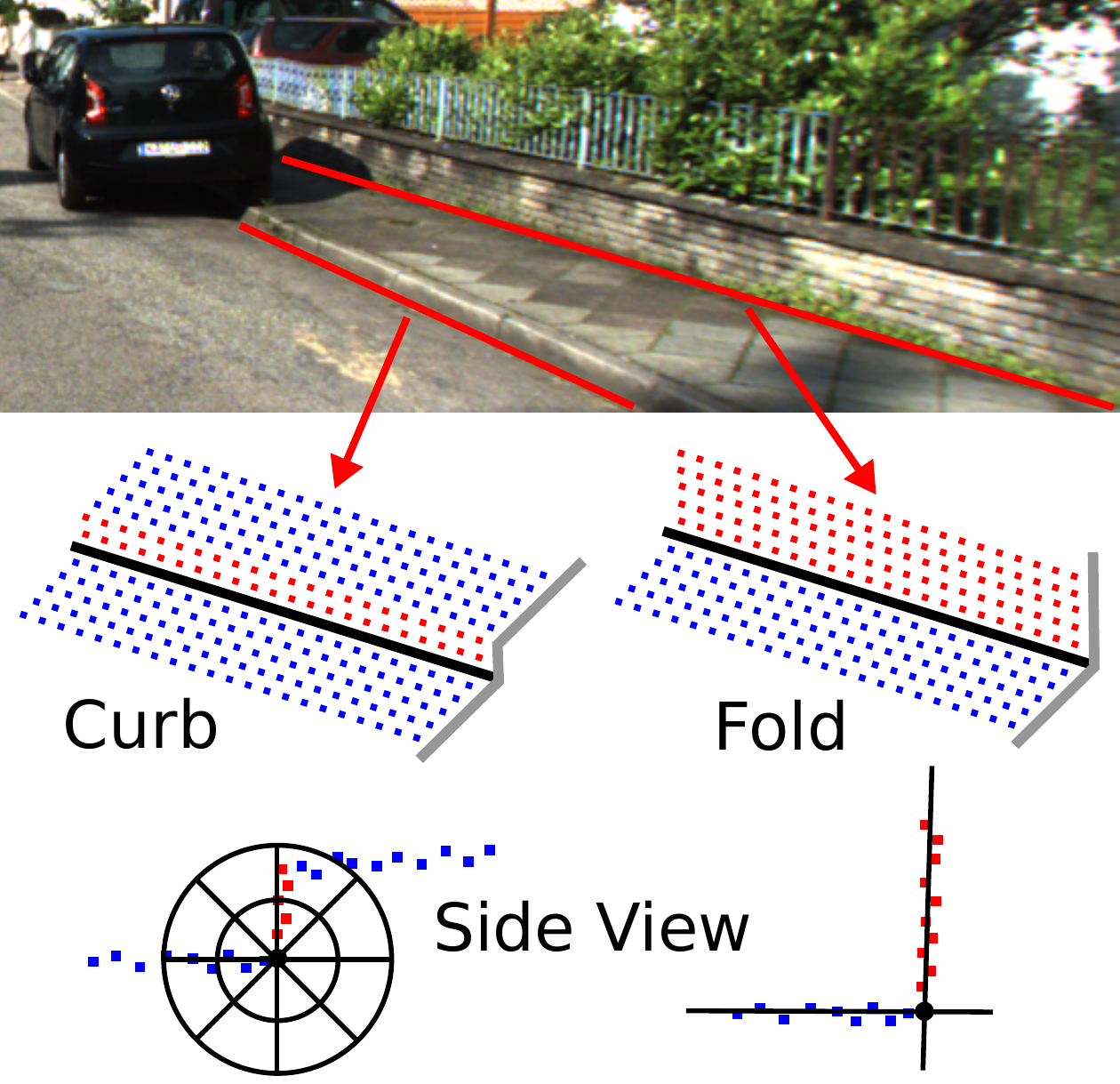}
\caption{3D Curbs and Folds}
\label{fig:model_fold_detection}
\end{subfigure}%
\caption{{\bf Label Transfer Model.} (\subref{fig:model_dense_crf}) Factor graph representation of our model.
(\subref{fig:model_fold_detection}) 3D structures such as folds and curbs are leveraged to improve segmentation boundaries between the categories  ``Road'', ``Sidewalk'' and ``Wall''.
}
\label{fig:model}
\vspace{-0.4cm}
\end{figure}

Given sparse point clouds and 3D annotations, we are interested in generating dense semantic instance annotations for all images.
Towards this goal, we propose a CRF model which reasons jointly about the labels of the 3D points and all pixels in the image, leveraging the calibration and registration described in \secref{sec:data_collection}.
Note that our 3D annotations are sparse and noisy, \ie, 3D points can carry none, one or multiple labels due to overlapping bounding primitives in 3D. The algorithm described in this section is designed to resolve these situations and infers marginal estimates for all 3D points and pixels in the image. In order to make our approach more robust in regions where appearance is not discriminative, we investigate additional geometric cues of the 3D point cloud such as 3D surface folds and curbs (see \figref{fig:model_fold_detection}). If detected, these cues can provide accurate boundaries between semantic classes in the image.

More formally, let $\cP$, $\cL$ and $\cF$ denote the set of image pixels, sparse 3D points from laser/stereo,
and detected 3D fold or curb segments, respectively. For each pixel $i\in\cP$ and each 3D point $l\in\cL$, we specify random variables $s_i$ and $s_l$ taking values from the set of semantic (or instance) labels $\{1,\dots,S\}$, where $S$ denotes the number of classes.
For instance inference, we assign a unique ID to each object which projects into the image. Thus, semantic and instance inference can be treated equally under our model and we will refer to both as ``semantic labels'' in the following.

Let $\bs = \{s_i|i\in\cP\} \cup \{s_l|l\in\cL\}$ denote the set of semantic labels. Dropping all dependencies on the image and point cloud for clarity we specify our CRF in terms of the following Gibbs energy function:
\begin{align}
&E(\bs) = \sum_{i \in \cP} \varphi_{i}^{\cP}(s_i)
+ \sum_{l \in \cL} \varphi_{l}^{\cL}(s_l)
+ \hspace{-0.1cm} \sum_{m\in\cF} \sum_{i\in\cP}  \varphi_{mi}^{\cF}(s_i) \label{eq:crf_energy}\\
&+ \hspace{-0.1cm} \sum_{i,j\in\cP} \hspace{-0.1cm}\psi_{ij}^{\cP,\cP}(s_i,s_j)
+ \hspace{-0.1cm} \sum_{l,k\in\cL} \hspace{-0.1cm} \psi_{lk}^{\cL,\cL}(s_l,s_k) 
+ \hspace{-0.3cm} \sum_{i \in \cP, l \in \cL} \hspace{-0.2cm} \psi_{il}^{\cP, \cL}(s_i,s_l)\nonumber
\end{align}
with unary potentials $\varphi(\cdot)$ and pairwise potentials $\psi(\cdot)$. For notational clarity, we omit all conditional dependencies on the input images, 3D points and 3D annotations.

\noindent{\bf Pixel Unary Potentials:}
The pixel unary potentials $\varphi_i^{\cP}(s_i)$ encode the likelihood of pixel $i$ taking label $s_i$
\begin{equation}
\varphi_i^{\cP}(s_i) =  w^{\cP}_1(s_i) \, \xi^{\cP}_i(s_i) -\,w^{\cP}_2(s_i) \log p^{\cP}_i(s_i)
\end{equation}
where $w^\cP_1$ and $w^\cP_2$ denote learned feature weights.
Our first constraint $\xi^{\cP}_i(s_i)$ determines the set of admissible labels and is obtained by projecting the 3D bounding primitives (which are an upper bound on the objects' extent) into the image.
We formulate the constraint via a binary feature $\xi^{\cP}_i(s_i)\in\{0,1\}$ which takes $0$ for pixel $i$ if its ray passes through a primitive of class $s_i$, and $1$ otherwise.

In addition, we leverage appearance information by projecting all non-occluded sparse 3D points into all adjacent frames of the image sequence and training a pixel-wise classifier \cite{Shotton2009IJCV} based on these projections. This results in a per-pixel probability distribution over semantic labels $p^{\cP}_i(s_i)$.
The intuition behind this feature is that regions of the same semantic class are similar in adjacent frames and thus yield highly discriminative cues for the current frame.

\noindent{\bf 3D Point Unary Potentials:}
The 3D point unary potentials $\varphi_l^{\cL}(s_l)$ encode the likelihood of 3D point $l$ taking label $s_l$:
\begin{equation}
\varphi_{l}^{\cL}(s_l) = -w^{\cL}(s_l)  \, \xi^{\cL}_l(s_l)
\end{equation}
where $\xi^{\cL}_l(s_l)$ denotes a feature which takes $0$ if the 3D point $l$ lies within a 3D primitive of class $s_l$, and $1$ otherwise. As the ``sky'' class can't be modeled with primitives we set $\xi^{\cL}_l(s_l)$ to $0$ if $s_l$ takes the label ``sky''. Additionally, we create ``virtual sky points'' at infinity for all pixels whose ray doesn't intersect any 3D primitive. Note that these pixels must correspond to sky regions as we assume that each object is completely contained in one or several bounding 3D primitive(s).

\begin{figure}[t]
\includegraphics[width=\linewidth]{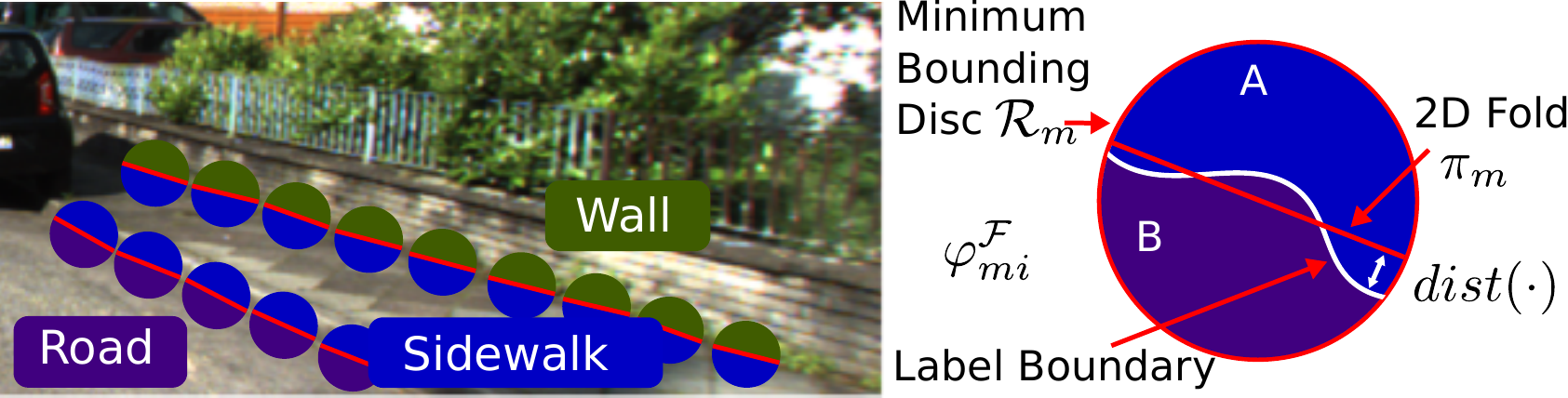}
\caption{{\bf Geometric Unary Potentials.} Left: We encourage label changes at 3D curbs or folds after projection into the image domain. Right: This constraint ($\varphi_{mi}^{\cF}$) is implemented by pixel unary potentials inside each minimum bounding disc $\cR_m$ around each 2D curb or fold segment $m$.}
\label{fig:geometric_unary_potentials}
\vspace{-0.4cm}
\end{figure}

\noindent{\bf Geometric Unary Potentials:}
We encourage label changes at curbs or folds which we detect in 3D using plane fitting as described in the supplementary document.
Given the projections into 2D, we introduce the following constraint:
\begin{equation}
\varphi_{mi}^{\cF}(s_i) = w^{\cF} \, \frac{[\bp_i\in\cR_m \land \nu_m(\bp_i) \neq s_i]}{\exp \left\{dist(\bp_i,\bpi_m) \right\}}
\label{eq:fold_consistency}
\end{equation}
Here, $[\cdot]$ is the Iverson bracket, $\bp_i$ denotes the 2D location of pixel $i$ and $\cR_m$ represents a 2D disc around curb or fold segment $m$ projected into 2D (yielding a line segment $\bpi_m$) as illustrated in \figref{fig:geometric_unary_potentials}. $\nu_k$ is a function which takes as input a pixel location and returns the semantic label predicted by fold $m$. More specifically, we project the 3D fold into 2D and compute the majority label at its two sides from the sparse projected 3D points. The denominator in \eqref{eq:fold_consistency} ensures a penalty decay towards the disc boundaries.

\noindent{\bf Pixel Pairwise Potentials:}
Our dense pairwise term encourages semantic label coherence and connects all pixels in the image via Gaussian edge kernels
\begin{eqnarray}
&&\psi_{ij}^{\cP,\cP}(s_i,s_j) ~=~
w_1^{\cP,\cP}(s_i,s_j)\exp\left\{-\frac{\lVert  \bp_{i} - \bp_{j} \rVert^2}
{2\,\theta_1^{\cP,\cP}}\right\} \nonumber\\
&&+\,
w_2^{\cP,\cP}(s_i,s_j)\exp\left\{-\frac{\lVert  \bp_{i} - \bp_{j} \rVert^2}
{2\,\theta_2^{\cP,\cP}}-\frac{\lVert  \bc_{i} - \bc_{j} \rVert^2}
{2\,\theta_3^{\cP,\cP}}\right\}\hspace{0.6cm}
\end{eqnarray}
where $\bp_i$ is the 2D location of pixel $i$ and $\bc_i$ denotes its color value.
Further, $w_1^{\cP,\cP}$ and $w_2^{\cP,\cP}$ are learned pairwise feature weights and $\theta^{\cP,\cP}$ parametrizes the kernel width.

\noindent{\bf 3D Pairwise Potentials:}
Similarly, we apply a Gaussian edge kernel to encourage label consistency between 3D points based on their 3D location and surface normals
\begin{eqnarray}
\psi_{lk}^{\cL, \cL}(s_l,s_k) &=& w^{\cL,\cL}(s_l,s_k)\\
&\times&\exp\left\{-\frac{\lVert \bp_{l}^{3d} - \bp_{k}^{3d} \rVert^2}
{2\,\theta_1^{\cL,\cL}}-\frac{{(n_{l} - n_{k})}^2}
{2\,\theta_2^{\cL,\cL}}\right\}\nonumber
\end{eqnarray}
where $\bp^{3d}_l$ is the 3D location of point $l$ and $n_l$ denotes the vertical (up) component of its normal. We use the normal's z-component as it is the most discriminative cue for indicating label changes between horizontal (\eg, road, sidewalk) and vertical (\eg, side of car, wall) surfaces. We estimate the respective normals using principle component analysis in a local neighborhood around each 3D point.

\noindent{\bf 2D/3D Pairwise Potentials:}
Finally, we encourage coherence between all 3D points and the image pixels
\begin{equation}
\psi_{il}^{\cP, \cL}(s_i,s_l) =
w^{\cP, \cL}(s_i,s_l)\exp\left\{-\frac{\lVert \bp_{i} -\bpi_l \rVert^2}
{2\,\theta^{\cP, \cL}}\right\}
\end{equation}
where $\bpi_l$ denotes the projection of the 3D laser or stereo point $l$ onto the image plane. Importantly, we project only points into the image which are likely to be visible. We determine these points by meshing the 3D point cloud using the ball-pivoting method of Bernardini \etal \cite{Bernardini1999VCG}, considering only 3D points in front of the mesh. We also tried state-of-the-art multi-view reconstruction approaches \cite{Jancosek2011CVPR} for mesh generation, but obtained better results with the described meshing approach.

\subsection{Learning and Inference}
\label{sec:learning_inference}

This section describes inference and parameter estimation in our label transfer model.

\noindent{\bf Inference:}
At test time, we are interested in estimating the marginal distribution of each semantic or instance label in $\bs$ under our model, specified by the Gibbs distribution defined in \eqref{eq:crf_energy}. The most likely configuration can then be estimated by variable-wise maximization of these marginals.
As our graphical model is loopy, exact inference in polynomial time is
intractable. We resort to variational inference and approximate the
probability distribution on $\bs$ by replacing it with a factorized mean
field distribution $Q(\bs) = \prod_{i\in \cP\cup \cL} Q_i(s_i)$.
This mean field approximation can be  computed efficiently using bilateral filtering \cite{Kraehenbuehl2011NIPS}.
As our model comprises three sets of densely connected variables (namely $\cP$, $\cL$ and $\cP\leftrightarrow\cL$), we exploit the algorithm of \cite{Kiefel2014ECCV,Vineet2013EMMCVPR} which generalizes \cite{Kraehenbuehl2011NIPS} to multiple fields.

\noindent{\bf Learning:}
We employ empirical risk minimization in order to learn the parameters in our model, considering the univariate logistic loss, defined as
$\Delta(s) = -\log \left(P(s)\right)$
where $P(\cdot)$ denotes the marginal distribution at the respective site. Let
us subsume all model parameters into $\Theta = \{w^\cP_1, w^\cP_2, w^\cL,w^{\cF},w^{\cP,\cP}_1,w^{\cP,\cP}_2,w^{\cP,\cL},w^{\cL,\cL}\}$. We define our minimization objective $f(\Theta)$ as the regularized univariate logistic loss:
\begin{eqnarray}
f(\Theta) &=& \sum_{n=1}^N \, \sum_{i\in \cP} - \log \left(Q_{n,i}(s_{n,i}^*)\right) + \lambda \, C(\Theta)\hspace{0.5cm}
\end{eqnarray}
Here, $N$ is the number of training images, $s_{n,i}^*$ denotes the ground truth semantic label and $Q_{n,i}(\cdot)$ the approximate marginal at pixel $i$ in image $n$, calculated via mean field approximation. $C(\Theta)$ is a quadratic regularizer on the parameter vector $\Theta$. We whiten all features and use a single value $\lambda$ which we select via cross-validation on the training set.
For learning the instance segmentation parameters we exploit the same loss $f(\Theta)$ as for semantic segmentation. For instance segmentation, we assign unique labels to each individual object, e.g., different cars will be assigned different labels even if they occlude each other. In order to associate 2D ground truth instances with 3D instances we project all visible 3D points into the image and find a consensus via the majority vote which gave good results in practice. As the number of instances per semantic class varies between images, we learn intra- and inter-class pairwise potentials using parameter tying.
We optimize the objective function $f(\Theta)$ using stochastic gradient descent and obtain $\partial Q / \partial \Theta$ using auto differentiation. 
We make use of the ADADELTA algorithm \cite{Zeiler2013ARXIV} with decay parameter $0.95$ and $\epsilon=10^{-8}$, and randomly sample a batch of $16$ training images at each iteration for which all gradients can be computed in parallel.

\section{Experimental Evaluation}
\label{sec:results}

In this section, we first evaluate our method in ablation studies and with respect to several label transfer baselines. Besides, we exploit the uncertainty in our predictions to increase accuracy for semi-dense predictions. Finally, we show some qualitative results of our method.
As input to our method, we accumulate all laser measurements in a common world coordinate system and augment them with 3D points from stereo matching \cite{Hirschmueller2008PAMI}. To reduce outliers, we consider only points up to $15$ m distance, and apply left-right as well as forward-backward consistency checks over 5 frames. We fuse all 3D points into one global point cloud and remove all points which are closer than $5$ cm to their nearest neighbor.
For evaluation, we manually annotated $160$ images from $8$ different suburbs with dense pixel-wise ground truth. From the $160$ frames, $120$ frames have been labeled in equidistant steps of $5$ frames for comparison with 2D label transfer methods.
We learn the parameters in our and the baseline models using $2$-fold cross validation at the sequence level to avoid any bias caused by the correlation of adjacent frames within a sequence. The kernel width parameters in our model have been chosen empirically as detailed in our supplementary document.

\subsection{Quantitative Evaluation}
\label{sec:results_quantitative}

This section presents our quantitative evaluation. We compare our method with respect to several baselines on the semantic and instance segmentation tasks.

\noindent{\bf Semantic Segmentation:}
For evaluating semantic segmentation performance, we map the 27 semantic labels in our 3D annotations to the most frequently occurring 14 categories (see supplementary material).
We report the frequency of these classes in the supplementary material.
We measure overall performance by the average Jaccard Index (JI) weighted by the class frequency and the average pixel accuracy (Acc).

The upper half of \tabref{tab:baseline} shows results of several 2D to 2D label transfer methods on all $120$ equidistantly labeled frames. Here, the task is to predict the center frame from two annotated images ($\pm5$ frames corresponding to $0.5$ seconds of driving or $\sim5$ meters travel distance).
\begin{figure}[t!]
\begin{subfigure}{.48\linewidth}
\includegraphics[width=\linewidth]{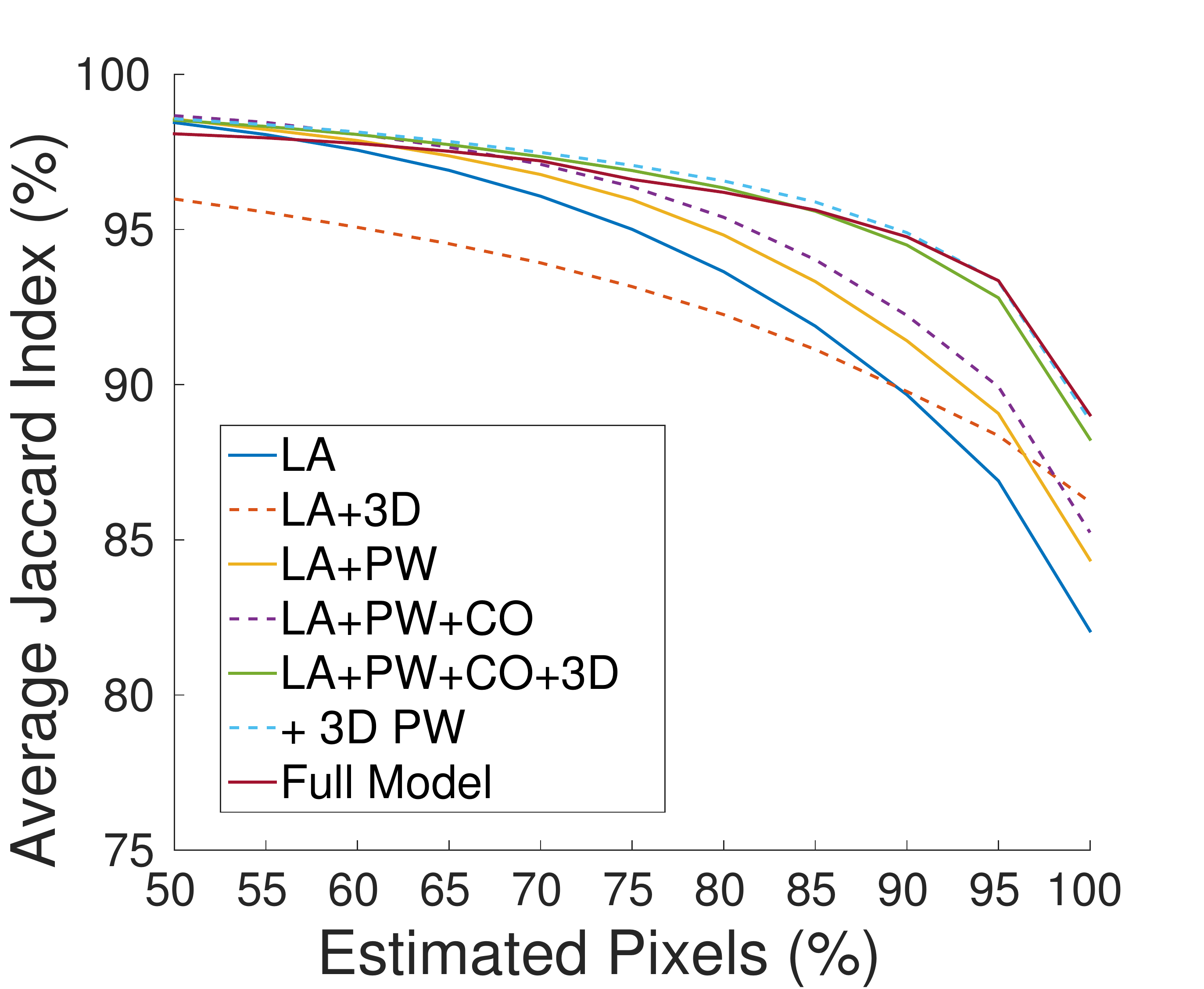}
\caption{Jaccard Index}
\label{fig:density_jaccard_index}
\end{subfigure}%
\hfill
\begin{subfigure}{.48\linewidth}
\includegraphics[width=\linewidth]{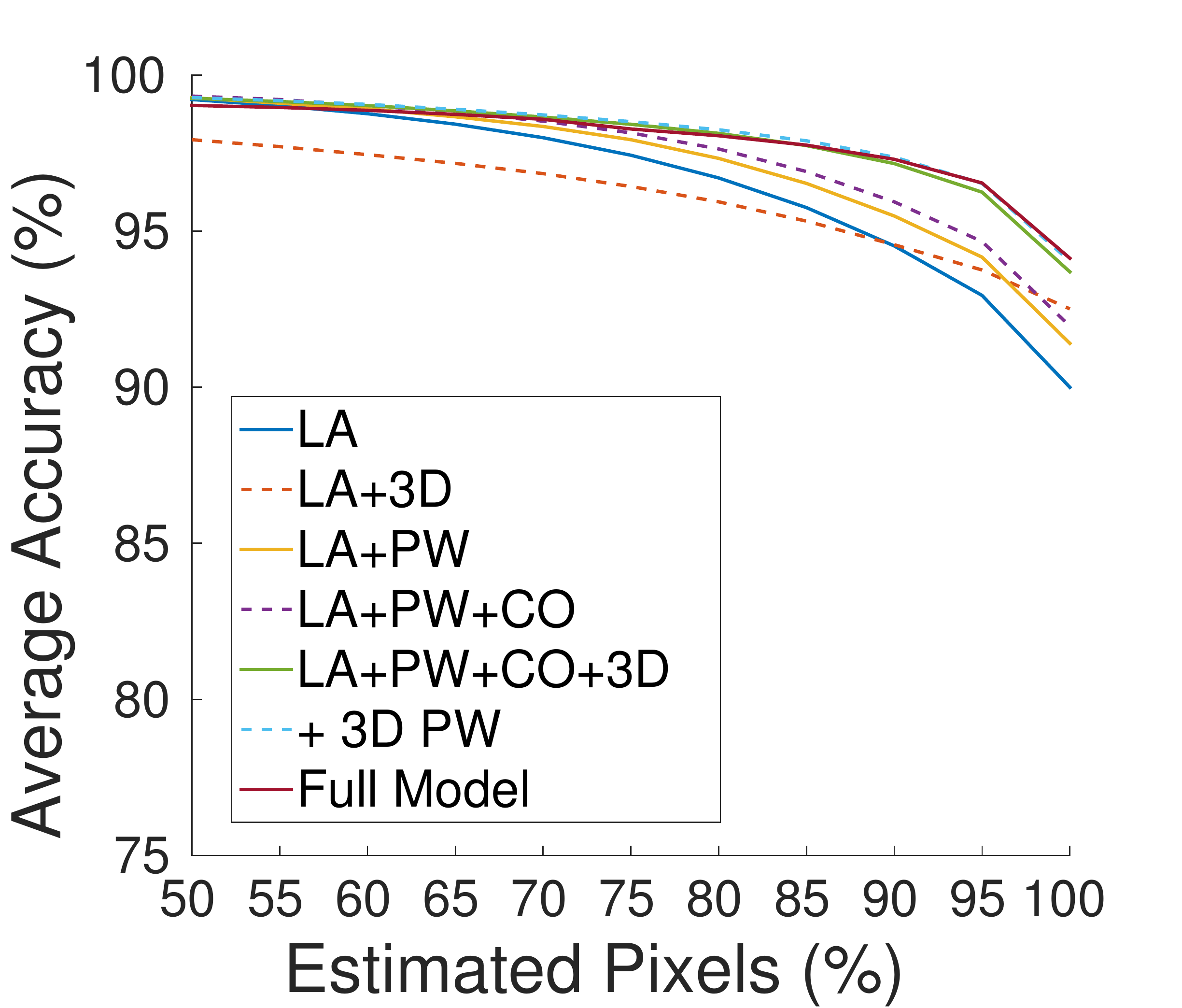}
\caption{Accuracy}
\label{fig:density_accuracy}
\end{subfigure}%
\vspace{-0.1cm}
\caption{{\bf Performance \wrt Estimated Pixels.} This figure shows the average Jaccard Index (\subref{fig:density_jaccard_index}) and the average accuracy (\subref{fig:density_accuracy}) when estimating only a fraction of the pixels which is selected according to the uncertainty in our predictions.}
\label{fig:density}
\vspace{-0.5cm}
\end{figure}
Our first baseline (``Label Prop.'') is the label transfer approach presented in \cite{Vijayanarasimhan2012ECCV}. To ensure that all baselines have access to the same information, we do not select frames in an active fashion but use equidistantly spaced labels for all methods (the driving speed during recording was nearly constant).
We construct a second baseline (``Sparse Track. + GC'') using the feature tracking approach of \cite{Sundaram2010ECCV} to propagate semantic labels from the two closest labeled frames to the target frame. To densify the label map, we apply graph cuts (GC) with contrast sensitive edge potentials \cite{Boykov2004PAMI}.

In order to evaluate the value of 3D information, we implemented a third baseline (``3D Prop. + GC'') which works similar to the previous one, but replaces the sparse tracking part with correspondences obtained by transferring pixels of the two closest labeled frames to the target image via the visible vertices of our 3D mesh followed by graph cuts propagation.
Finally, we train the segmentation model of Krähenbühl \etal \cite{Kraehenbuehl2011NIPS} (``Fully Conn. CRF'') on all annotated adjacent frames of the {\it test sequence}.

\begin{table*}[t!]
\setlength{\tabcolsep}{2.27pt}
\begin{tabular}{|p{3.5cm}|>{\centering}p{0.7cm}>{\centering}p{0.7cm}>{\centering}p{0.7cm}>{\centering}p{0.7cm}>{\centering}p{0.7cm}>{\centering}p{0.7cm}>{\centering}p{0.7cm}>{\centering}p{0.7cm}>{\centering}p{0.7cm}>{\centering}p{0.7cm}>{\centering}p{0.7cm}>{\centering}p{0.7cm}>{\centering}p{0.7cm}>{\centering}p{0.7cm}|>{\centering}p{0.7cm}|>{\centering}p{0.7cm}|}  \hline
\small Method & \small Road & \small Park & \small Sdwlk & \small Terr & \small Bldg & \small Vegt & \small Car & \small Trler & \small Carvn & \small Gate & \small Wall & \small Fence & \small Box & \small Sky & \small JI & \small Acc \tabularnewline \hline
{\small Label Prop. \cite{Vijayanarasimhan2012ECCV}}& 93.4 & 51.8 & 73.5 & 58.3 & 80.2 & 69.9 & 61.5 & 22.4 & 42.3 & 30.6 & 45.3 & 45.7 & 32.5 & 89.6 & 74.4 & 84.4 \tabularnewline 
{\small Sparse Track. + GC \cite{Sundaram2010ECCV}}& 89.6 & 37.1 & 69.0 & 54.2 & 84.6 & 79.5 & 78.2 & 2.5 & 35.3 & 3.2 & 38.9 & 32.9 & 7.0 & 91.0 & 77.8 & 87.3 \tabularnewline 
{\small 3D Prop. + GC}& 91.3 & 44.5 & 74.0 & 62.4 & 86.2 & 81.8 & 81.6 & 5.2 & 38.6 & 12.7 & 47.4 & 42.0 & 15.0 & 88.8 & 80.2 & 88.9 \tabularnewline 
{\small Fully Conn. CRF \cite{Kraehenbuehl2011NIPS}}& 88.5 & 37.8 & 68.4 & 55.8 & 85.5 & 79.8 & 76.8 & 2.5 & 30.6 & 2.9 & 38.3 & 32.4 & 0.0 & \bf 92.8& 77.9 & 87.4 \tabularnewline 
\hline 
{\small 3D Primitives + GC}& 78.7 & 46.4 & 43.9 & 46.5 & 54.9 & 55.4 & 55.1 & 72.3 & 54.6 & 51.0 & 40.2 & 52.3 & 40.2 & 55.4 & 56.4 & 72.1 \tabularnewline 
{\small 3D Mesh + GC}& 92.1 & 66.4 & 72.4 & 66.1 & 69.1 & 74.9 & 87.7 & 88.9 & 88.5 & 61.9 & 51.4 & 60.7 & 30.4 & 46.4 & 72.5 & 82.6 \tabularnewline 
{\small 3D Points + GC}& 93.1 & 72.4 & 78.9 & 72.4 & 81.9 & 77.2 & 88.1 & 92.4 & 91.1 & 70.2 & 66.8 & 68.7 & 62.4 & 69.9 & 80.5 & 89.0 \tabularnewline 
{\small Proposed Method}& \bf 95.3& \bf 80.6& \bf 86.4& \bf 81.0& \bf 90.9& \bf 86.9& \bf 91.5& \bf 94.9& \bf 91.8& \bf 73.6& \bf 78.9& \bf 79.4& \bf 73.0& 91.0 & \bf 89.2& \bf 94.2\tabularnewline 
\hline 
\end{tabular}

\vspace{-0.2cm}
\caption{{\bf Comparison to Label Transfer Baselines on Semantic Segmentation Task.} We compare our method to 2D label transfer baselines (top) and to 3D to 2D label transfer baselines (bottom) on $120$ consecutive images. See text for details.}
\label{tab:baseline}
\vspace{-0.1cm}
\end{table*}

\begin{table*}[t!]
\setlength{\tabcolsep}{2.65pt}
\begin{tabular}{|p{3.0cm}|>{\centering}p{0.7cm}>{\centering}p{0.7cm}>{\centering}p{0.7cm}>{\centering}p{0.7cm}>{\centering}p{0.7cm}>{\centering}p{0.7cm}>{\centering}p{0.7cm}>{\centering}p{0.7cm}>{\centering}p{0.7cm}>{\centering}p{0.7cm}>{\centering}p{0.7cm}>{\centering}p{0.7cm}>{\centering}p{0.7cm}>{\centering}p{0.7cm}|>{\centering}p{0.7cm}|>{\centering}p{0.7cm}|}  \hline
\small Method & \small Road & \small Park & \small Sdwlk & \small Terr & \small Bldg & \small Vegt & \small Car & \small Trler & \small Carvn & \small Gate & \small Wall & \small Fence & \small Box & \small Sky & \small JI & \small Acc \tabularnewline \hline
{\small LA}& 92.2 & 64.6 & 77.9 & 67.5 & 85.2 & 81.9 & 81.7 & 85.7 & 81.5 & 46.8 & 62.1 & 60.3 & 49.4 & 83.1 & 82.1 & 90.0 \tabularnewline 
{\small LA+3D}& 95.0 & 76.9 & 85.5 & 73.3 & 87.9 & 84.3 & 89.4 & 88.2 & 90.2 & 68.8 & 74.6 & 74.0 & 63.7 & 83.4 & 86.2 & 92.5 \tabularnewline 
{\small LA+PW}& 92.5 & 68.6 & 79.5 & 73.1 & 87.3 & 84.2 & 84.1 & 89.9 & 85.9 & 48.7 & 66.2 & 64.9 & 54.5 & 86.6 & 84.4 & 91.4 \tabularnewline 
{\small LA+PW+CO}& 93.0 & 72.7 & 81.2 & 73.8 & 87.7 & 84.5 & 85.7 & 90.9 & 88.4 & 57.7 & 70.4 & 69.6 & 57.6 & 86.9 & 85.2 & 92.0 \tabularnewline 
{\small LA+PW+CO+3D}& 93.2 & 78.6 & 85.0 & 76.3 & 90.6 & 86.7 & 89.1 & 90.9 & 92.7 & 68.5 & 77.8 & 78.9 & 67.8 & 90.7 & 88.2 & 93.7 \tabularnewline 
{\small + 3D PW}& 94.9 & 80.1 & 85.9 & 80.0 & 90.6 & 87.0 & 91.2 & 91.3 & 93.8 & 72.6 & 78.1 & 78.5 & 69.3 & 90.8 & 88.8 & 94.0 \tabularnewline 
{\small Full Model}& 95.4 & 80.1 & 87.1 & 80.0 & 90.6 & 87.0 & 91.2 & 91.3 & 93.9 & 72.6 & 78.4 & 78.6 & 69.4 & 90.8 & 89.0 & 94.1 \tabularnewline 
\hline 
{\small Full Model (90\%)}& 98.1 & 92.3 & 94.7 & 92.4 & 95.3 & 93.5 & 96.5 & 95.8 & 97.6 & 83.7 & 90.7 & 90.7 & 84.0 & 94.6 & 94.9 & 97.4 \tabularnewline 
{\small Full Model (80\%)}& 98.8 & 95.3 & 96.7 & 94.9 & 96.8 & 95.5 & 97.5 & 96.4 & 98.5 & 86.4 & 93.7 & 93.4 & 87.9 & 96.4 & 96.6 & 98.2 \tabularnewline 
{\small Full Model (70\%)}& \bf 99.2& \bf 96.8& \bf 97.9& \bf 96.4& \bf 97.5& \bf 96.8& \bf 97.9& \bf 97.2& \bf 99.0& \bf 88.1& \bf 95.0& \bf 94.6& \bf 90.1& \bf 97.2& \bf 97.5& \bf 98.7\tabularnewline 
\hline 
\end{tabular}

\vspace{-0.2cm}
\caption{{\bf Ablation Study on Semantic Segmentation Task.} This table shows the importance of the different components in our model on all $160$ images. The components are abbreviated as follows:
LA = local appearance ($p^{\cP}$), PW = 2D pairwise constraints ($\psi^{\cP,\cP}$), CO = 3D primitive constraints ($\xi^{\cP}$), 3D = 3D points ($\varphi^{\cL}$,$\psi^{\cP, \cL}$), 3D PW = 3D pairwise constraints ($\psi^{\cL,\cL}$), Full Model = all potentials including folds. Percentages denote fractions of estimated pixels. See text for details.}
\label{tab:ablation_semantic}
\vspace{-0.1cm}
\end{table*}

\begin{table*}[t!]
\setlength{\tabcolsep}{2.65pt}
\begin{tabular}{|p{3.0cm}|>{\centering}p{0.7cm}>{\centering}p{0.7cm}>{\centering}p{0.7cm}>{\centering}p{0.7cm}>{\centering}p{0.7cm}>{\centering}p{0.7cm}>{\centering}p{0.7cm}>{\centering}p{0.7cm}>{\centering}p{0.7cm}>{\centering}p{0.7cm}>{\centering}p{0.7cm}>{\centering}p{0.7cm}>{\centering}p{0.7cm}>{\centering}p{0.7cm}|>{\centering}p{0.7cm}|>{\centering}p{0.7cm}|}  \hline
\small Method & \small Road & \small Park & \small Sdwlk & \small Terr & \small Bldg & \small Vegt & \small Car & \small Trler & \small Carvn & \small Gate & \small Wall & \small Fence & \small Box & \small Sky & \small JI & \small Acc \tabularnewline \hline
{\small LA+3D}& 94.5 & 74.7 & 83.5 & 73.4 & 80.7 & 84.5 & 86.3 & 90.8 & 90.9 & 66.3 & 74.7 & 75.6 & 63.1 & 81.9 & 83.5 & 91.0 \tabularnewline 
{\small LA+PW+CO}& 92.8 & 70.3 & 79.8 & 73.9 & 64.9 & 84.6 & 82.2 & 90.7 & 87.1 & 51.7 & 67.8 & 66.6 & 24.7 & 88.0 & 78.4 & 87.4 \tabularnewline 
{\small LA+PW+CO+3D}& 94.6 & 78.4 & 84.2 & 78.4 & 86.3 & 87.6 & 90.8 & 93.0 & 93.3 & 70.9 & 77.6 & 79.4 & 68.6 & 91.1 & 87.5 & 93.3 \tabularnewline 
{\small + 3D PW}& 95.1 & \bf 80.6 & 85.3 & \bf 79.3 & 86.4 & \bf 87.9 & 91.5 & 93.0 & 93.6 & \bf 73.6 & 78.1 & 79.0 & 70.4 & 90.7 & 87.9 & 93.5 \tabularnewline 
{\small Full Model}& \bf 95.7 & \bf 80.6 & \bf 86.9 & 79.2 & \bf 86.4 & \bf 87.9 & \bf 91.5 & \bf 93.1 & \bf 93.6& \bf 73.6 & \bf 78.5 & \bf 79.1 & \bf 70.5 & \bf 90.7 & \bf 88.1 & \bf 93.6 \tabularnewline 
\hline 
\end{tabular}

\vspace{-0.2cm}
\caption{{\bf Ablation Study on Instance Segmentation Task} using the same abbreviations as in \tabref{tab:ablation_semantic}. See text for details.\hfill~}
\label{tab:ablation_instance}
\vspace{-0.3cm}
\end{table*}

\begin{figure*}[t]
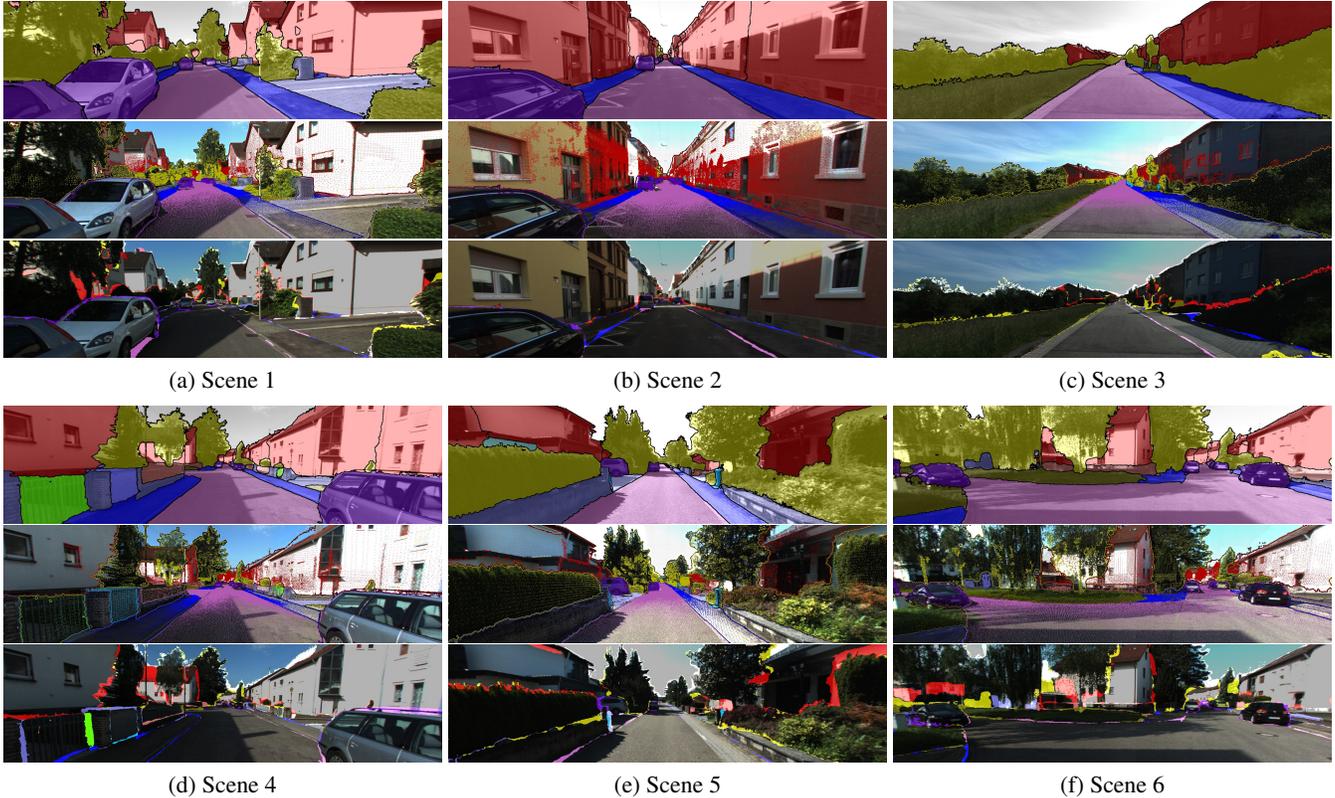

\def\qualwidth{0.333}
\def\qualmargin{0.1}
\newcommand{\qualresult}[3]{%
\begin{minipage}{#1\linewidth}%
\includegraphics[width=\linewidth]{results/instance/overlay/#2.jpg}\\%
\includegraphics[width=\linewidth]{results/semantics/pts/#2.jpg}\\%
\includegraphics[width=\linewidth]{results/semantics/error2/#2.jpg}\\%
\vspace{-0.5cm}%
\subcaption{#3}
\vspace{0.1cm}
\end{minipage}%
\hspace{\qualmargin cm}%
}
\qualresult{\qualwidth}{result_2013_05_28_drive_0000_sync_001930}{Scene 1}%
\qualresult{\qualwidth}{result_2013_05_28_drive_0000_sync_006450}{Scene 2}%
\qualresult{\qualwidth}{result_2013_05_28_drive_0004_sync_001475}{Scene 3}\\%
\qualresult{\qualwidth}{result_2013_05_28_drive_0000_sync_001750}{Scene 4}%
\qualresult{\qualwidth}{result_2013_05_28_drive_0004_sync_006920}{Scene 5}%
\qualresult{\qualwidth}{result_2013_05_28_drive_0000_sync_001780}{Scene 6}%
\vspace{-0.3cm}
\caption{{\bf Qualitative Results.} Each subfigure shows from top-to-bottom: the input image with inferred semantic instance segmentation, the projected 3D points and inferred semantic segmentation boundaries, as well as the errors with respect to 2D ground truth annotation where colors indicate ground truth labels. See supplementary material and text for details.}
\label{fig:qualitative_results}
\vspace{-0.4cm}
\end{figure*}

From the 2D label transfer baselines, the mesh transfer method which uses projected 3D information performs best. Furthermore, and maybe surprisingly, the image-specific fully connected CRF model performs on par or even better than special purpose label transfer methods. According to our experiments, this is caused by the fact that optical flow (as used in \cite{Vijayanarasimhan2012ECCV,Sundaram2010ECCV}) often fails for street scenes like ours due to large displacements, perspective distortions, textureless regions and challenging lighting conditions. On the other hand, the fully connected model performs weaker for less frequent or textureless classes such as ``Trailer'' or ``Box''.

The bottom half of \tabref{tab:baseline} compares the proposed method with respect to several 3D to 2D label transfer baselines which in contrast to the 2D to 2D label transfer methods exploit our 3D annotations and don't require equidistantly labeled 2D annotations.
As evidenced by our results, simply projecting 3D primitives or meshes into the image and smoothing via GC does not perform well due to the crude approximation of the geometry (``3D Primitives + GC''; ``3D Mesh + GC''). Better results are obtained when projecting the visible 3D points followed by spatial propagation (``3D Points + GC'').

Finally, we observe that all baselines are outperformed by the proposed method (last row) in almost all categories. Importantly, note that the 2D methods require every 10th frame to be labeled, while our method (as well as the other 3D baselines) require 3D annotations in the form of 3D primitives. Assuming $60$ minutes annotation time per image, this amounts to $20$ hours of annotation time per batch of $200$ frames when labeling one 2D image every $10$th frame, while the respective 3D annotations for this scene can be obtained in less than $3$ hours. Note that labeling each frame of the sequence manually would require $200$ hours. This gain multiplies with the frame rate and the number of cameras (our setup comprises four).

\noindent{\it Ablation Study:}
We evaluate the importance of the individual components of our model in \tabref{tab:ablation_semantic} (top). Starting with the appearance classifier trained on the projected sparse 3D points ($p^{\cP}$), we incrementally add the terms related to the 3D points ($\varphi^{\cL}$,$\psi^{\cP, \cL}$), the semantic pairwise term between pixels ($\psi^{\cP,\cP}$), the 3D primitive constraints ($\xi^{\cP}$), the 3D pairwise constraints ($\psi^{\cL,\cL}$) and finally the remaining terms ($\varphi_{mi}^{\cF}$) as specified in \eqref{eq:crf_energy}.
We note that each component is able to increase performance.
As expected, we obtain the largest improvement by reasoning about the relationship between points in 3D and pixels in the image.
Integrating 3D fold and curb detections improves road boundaries slightly.

\noindent{\it Semi-dense Inference:}
Often, it is not necessary to label all pixels in every image for training a semantic segmentation model. In this section, we therefore leverage our model's awareness of label uncertainty to estimate semi-dense label maps with high accuracy. To quantify uncertainty, we measure the entropy of the label marginal distribution at every pixel. Sorting all pixels according to their entropy allows us to predict the most certain regions in the image. \tabref{tab:ablation_semantic} (bottom) and \figref{fig:density} show our results when predicting only those parts of the image. Note how this helps to boost our performance to $94.9\%$ JI and $97.4\%$ accuracy when predicting at $90\%$ pixel density. In contrast, uncertainty is not directly accessible in most of the baseline models as they are deterministic or rely on MAP estimates. The only exception is the ``Fully Conn. CRF'' baseline. We provide the corresponding experiment in the supplementary material.

\noindent{\bf Instance Segmentation:}
As time consistent 2D instance ground truth is hard to obtain, most existing 2D label transfer methods focus on the semantic segmentation problem. Therefore, we chose to evaluate instance segmentation performance in an ablation study. We annotated the classes ``Building'', ``Car'', ``Trailer'', ``Caravan'' and ``Box'' with instances in our 2D ground truth. While the remaining classes (\eg, ``Road'', ``Sky'') do not admit unambiguous instance labels,
we also report their performance as our model reasons about all instance and semantic classes jointly. \tabref{tab:ablation_instance} shows our results. Note how the instance segmentation results are on par with the semantic segmentations, demonstrating our model's intra-class separation ability.
Semi-dense instance results are provided in the supplementary.

\subsection{Qualitative Evaluation}
\label{sec:results_qualitative}

\figref{fig:qualitative_results} illustrates our dense inference results qualitatively for 6 different scenes in terms of semantic instance segmentation. The last row shows the error maps where colors indicate the true label (see supplementary for color coding).
While the proposed method is able to delineate most object boundaries satisfyingly, some challenges remain. Errors occur in low-contrast image regions with overlapping 3D annotations (scene 1: car/road boundary) and in regions where 3D points are absent due to sensor occlusion (scene 4: building roof). Another source of errors are inherent label ambiguities which occur for porous objects such as fences or trees (scene 6: tree boundary) where even 2D ground truth annotation is a hard and ambiguous task. Finally, also manual 2D annotations contain errors, in particular at complex boundaries which are hard to delineate (scene 4: trees, scene 5: hedge). However, note that our semi-dense inference is able to successfully identify those regions as shown in \figref{fig:density} and our supplementary material.

\section{Conclusion}

We presented a method for semantic instance labeling of large datasets from annotated 3D primitives. In the presence of 3D data, our method yields better results compared to several state-of-the-art 2D label transfer baselines while lowering annotation time. Furthermore, our method yields temporally consistent instance labels and explicitly exposes label uncertainty. We also proposed a novel dataset comprising 400k images, laser point clouds and annotations for all objects which we make publicly available. In future work, we plan to extend our method to dynamic scenes by joint inference over multiple frames.

{\small
\bibliographystyle{ieee}
\bibliography{bibliography_long,bibliography}
}

\end{document}